\documentclass[11pt,a4paper]{article}
\usepackage[hyperref]{naaclhlt2019}
\usepackage{float}
\usepackage{times}
\usepackage{latexsym}
\usepackage{amsmath}
\usepackage{amsfonts}
\usepackage{csvsimple}
\usepackage{url}
\usepackage{graphicx}  
\usepackage{subfigure}
\usepackage{multirow}

\aclfinalcopy 

\title{Ranking-Based Autoencoder for Extreme Multi-label Classification}

\author{Bingyu Wang\\ 
Khoury College of CS \\
Northeastern University \\
{\tt rainicy@ccs.neu.edu} \\\And
Li Chen \\
JD.com Inc. \\
{\tt li.chen@jd.com} \\\And
Wei Sun \\
Department of ECE \\
North Carolina State University \\
{\tt wsun12@ncsu.edu} \\\AND
Kechen Qin \\
Khoury College of CS \\
Northeastern University \\
{\tt qin.ke@husky.neu.edu} \\\And
Kefeng Li \\
JD.com Inc. \\
{\tt likefeng@jd.com} \\\And
Hui Zhou \\
JD.com Inc. \\
{\tt hui.zhou@jd.com}}

\date{}
\begin{document}
\maketitle
\begin{abstract}
Extreme Multi-label classification (XML) is an important yet challenging machine learning task, that assigns to each instance its most relevant candidate labels from an extremely large label collection, where the numbers of labels, features and instances could be thousands or millions. XML is more and more on demand in the Internet industries, accompanied with the increasing business scale / scope and data accumulation. The extremely large label collections yield challenges such as computational complexity, inter-label dependency and noisy labeling. Many methods have been proposed to tackle these challenges, based on different mathematical formulations. In this paper, we propose a deep learning XML method, with a word-vector-based self-attention, followed by a ranking-based AutoEncoder architecture. The proposed method has three major advantages: 1) the autoencoder simultaneously considers the inter-label dependencies and the feature-label dependencies, by projecting labels and features onto a common embedding space; 2) the ranking loss not only improves the training efficiency and accuracy but also can be extended to handle noisy labeled data; 3) the efficient attention mechanism improves feature representation by highlighting feature importance. Experimental results on benchmark datasets show the proposed method is competitive to state-of-the-art methods.
\end{abstract}

\section{Introduction and Related Work}

In multi-label classification~\cite{tsoumakas2007multi,zhang2014review}, one  assigns multiple labels to each instance. Multi-label classification has many real-word applications: for example, a movie may be associated with multiple genres, a web page may contain several topics, and an image can be tagged with a few objects. In these classification tasks, labels often exhibit complex dependencies: for example, \textit{Documentary} and \textit{Sci-Fi} are usually mutually exclusive movie genres, while \textit{Horror} and \textit{Thriller} are typically highly correlated. Predicting labels independently fails to capture these dependencies and suffers suboptimal performance~\cite{tsoumakas2007multi,ghamrawi2005collective,li2016conditional}. 
%
%
%
Several methods that capture label dependencies have been proposed, including Conditional Random Fields (CRF)~\cite{lafferty2001conditional,ghamrawi2005collective}, Classifier Chains (CC)~\cite{read2011classifier,dembczynski2010bayes}, Conditional Bernoulli Mixtures (CBM)~\cite{li2016conditional}, and Canonical Correlated AutoEncoder (C2AE)~\cite{yeh2017learning}. 
However, these methods typically only work well on small-to-medium scale datasets.

Extreme multi-label classification (XML) is a multi-label classification task in which the number of instances, features and labels are very large, often on the order of thousands to millions~\cite{zubiaga2012enhancing,bhatia2015sparse}. It has numerous real-world applications such as merchandise tagging and text categorization. Although the label vocabulary is large, typically each instance only matches a few labels. The scale of the classification task, the inter-dependent labels, and label sparsity all pose significant challenges for accurate and efficient classification.


Many methods have been proposed for extreme multi-label classification. We group them into different categories and describe representative methods in each category.

\smallskip
\noindent
\textbf{Independent Classification:}
A popular method is to divide the multi-label classification problem into multiple binary classification problems~\cite{tsoumakas2007multi,hariharan2012efficient,babbar2017dismec,yen2016pd,yen2017ppdsparse}. A typical implementation is to treat labels independently and train one-vs-all classifiers for each of the labels. These independent classifiers can be trained in parallel and thus are computationally efficient in practice. Ignoring the inter-label dependency also enables efficient optimization algorithm, which further reduces computational cost. However, ignoring label dependency inherently limits prediction accuracy. A competitive method in this category is called PD-Sparse~\cite{yen2016pd}, with a variant of the Block-Coordinate Frank-Wolfe training algorithm that exploits data sparsity and achieves complexity sub-linear in the number of primal and dual variables. PD-Sparse~\cite{yen2016pd} shows better performance with less training and prediction time than 1-vs-all Logistic Regression or SVM on extreme multi-label datasets.


\noindent
\textbf{Tree Based Classifiers:}
Following the success of tree-based algorithms in binary classification problems, people also proposed tree-based algorithms for multi-label classification~\cite{agrawal2013multi,weston2013label,prabhu2014fastxml}, which achieve promising prediction accuracy. Similar to decision trees, these methods make classification decisions in each branch split. Different from decision trees, each split evaluates all features, instead of one, to make a decision. Also, each decision is for a subset of labels rather than one label. Finally, via ensembling and parallel implementation, trees can boost their prediction accuracy with practically affordable computational cost. Among these tree based classifiers, FastXML~\cite{prabhu2014fastxml} further optimizes an nDCG-based ranking loss function and achieves significantly higher accuracy than other peer methods. 

\noindent
\textbf{Embedding:}
A major difficulty of extreme multi-label classification is the large number of labels.  When labels are inter-dependent, one can attempt to find a lower dimensional latent label space from which one can fully reconstruct the original label space.  Over the past decade, many methods were proposed to find this latent label space. In early work, methods were proposed to linearly project the original label space into a lower-dimension space and reconstruct predictions from that space~\cite{tai2012multilabel,balasubramanian2012landmark}. However, there are two assumptions: (1) the label dependency is linear and (2) the label matrix is low-rank, which do not always hold, as reflected by the low prediction accuracy of these methods. To overcome the limitation of the linear assumption, different methods were proposed using non-linear embeddings, including kernels, sub-sampling~\cite{yu2014large}, feature-aware~\cite{lin2014multi,yeh2017learning} and pairwise distance preservation~\cite{bhatia2015sparse}. Among these methods, SLEEC~\cite{bhatia2015sparse} stands out for less training time and higher accuracies. SLEEC introduces a method for learning a small ensemble of local pairwise distance preserving embeddings which allows it to avoid the low-rank and linear-dependency assumption.

\noindent
\textbf{Deep Learning:}
Deep learning has not been well studied for XML, although it has achieved great successes in binary and multi-class classification problems~\cite{lin2017structured,kim2014convolutional}.



FastText~\cite{grave2017bag} reconstructs a document representation by averaging the embedding of the words in the document, followed by a softmax transformation. It is a simple but very effective and accurate multi-class text classifier, as demonstrated in both sentiment analysis and multi-class classification~\cite{grave2017bag}. However, FastText may not be directly applicable for more complicated problems, like XML.


BoW-CNN~\cite{johnson2014effective} learns powerful embedding of small text regions by applying CNN to high-dimensional text data. The embedding of all regions are sent to one or multiple convolutional layers, a pooling layer and the output layer at the end.


XML-CNN~\cite{liu2017deep} achieves computational efficiency by training a deep neural network with a hidden bottleneck layer much smaller than the output layer. However, this method has a few drawbacks. First, it is trained using the binary cross entropy loss. This loss tends to be sensitive to label noise, which is frequently observed in extreme multi-label data. Since the label vocabulary is large, it is quite common for human annotator to miss relevant tags. When the classifier's prediction (which might be correct) disagrees with the annotation, the cross entropy loss can potentially assign an unbounded penalty to the classifier during training procedure. The second issue is that because labels are trained independently as separate binary classification tasks, their prediction probabilities/scores may not be directly comparable. This is problematic because in many applications the requirement is to rank all labels according to their relevance, as opposed to making an independent binary decision on each label. The third defect is that XML-CNN requires raw documents as input since it adopts the CNN structure on top of sentences~\cite{kim2014convolutional}; this is problematic when datasets are given in other formats such as bag-of-words for text.
%

C2AE~\cite{yeh2017learning} uses a ranking loss as the training objective. But the ranking loss employed there needs to compare all (positive label, negative label) pairs, and therefore does not scale well to extreme data.
%
%
Furthermore, C2AE only takes the bag-of-words representation (one-hot encoding) as the input, which makes it harder to learn powerful representations from extreme multi-label dataset.
%

\noindent
\textbf{Our Contribution}
In this paper, we propose a new deep learning method to address extreme multi-label classification. Our contributions are as follows:

\begin{itemize}
\item Motivated by the recent success of attention techniques, we propose an efficient attention mechanism that can learn rich representations from any type of input features, including but not limited to bag-of-words, raw documents and images.

\item Inspired by C2AE, our proposed model projects both features and labels onto common latent spaces wherein correlations between features and labels are exploited. By decoding this latent space into the original label space in the prediction stage, the dependencies between labels are implicitly captured. 

\item We propose a margin-based ranking loss that is simultaneously more effective for extreme settings and more tolerant towards noisy labeling.  


\end{itemize}


\begin{figure}[t]
\centering
\includegraphics[width=1.\columnwidth]{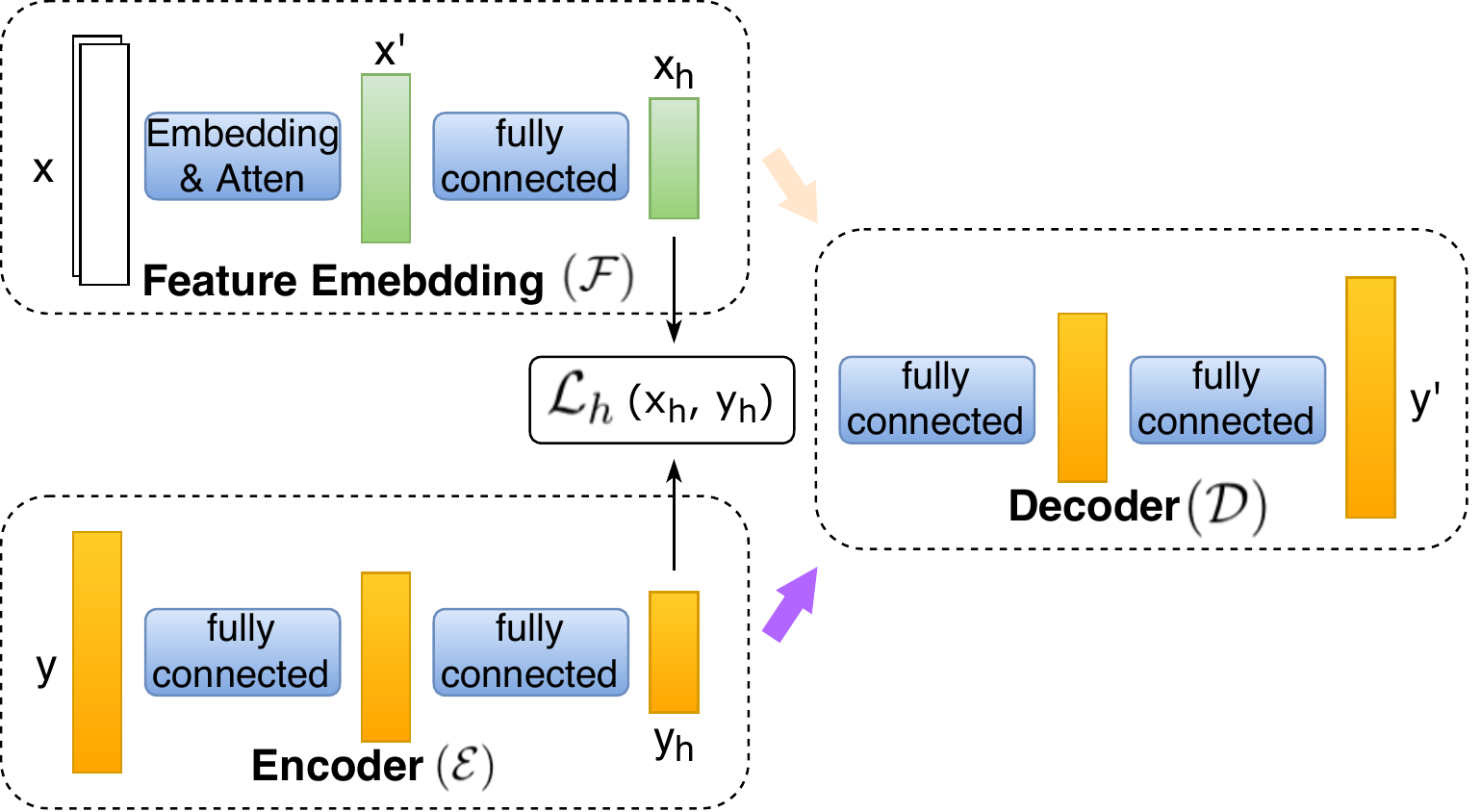}
\caption{Ranking-based AutoEncoder for XML.}
\label{fig:rank_ae}
\end{figure}

\section{The Proposed Method (Rank-AE)}
In this section, we introduce the data format in XML, and the proposed networks, including the marginal-based ranking loss and attentions. 
\subsection{Data Format}
In XML, we are given a set of label candidates $\mathcal{Y}=\{1,2,\dots,L\}$. The dataset $D$ consists of features and labels: $D = \{(x_i, y_i)\}_{i=1}^N$, wherein $N$ is number of data, and each instance $x\in\mathbb{R}^V$~($V$ is the feature dimension) matches a label subset $y \subseteq \mathcal{Y}$, which can be written as a binary vector $y = \{0,1\}^L$, with each bit $y_l$ representing the presence or absence of label $l$. Given such dataset, our goal is to build a classifier $c$:~$\mathbb{R}^V\rightarrow \{0,1\}^L$, mapping an instance to a subset of labels with arbitrary size.

\subsection{Auto-Encoder Network}\label{sec:ae_network}
Inspired by the C2AE~\cite{yeh2017learning}, we propose a Ranking-based Auto-Encoder (\textbf{Rank-AE}), as depicted in Figure~\ref{fig:rank_ae}. Similar to C2AE, Rank-AE includes three mapping functions to be trained: a mapping from input features $x$ to feature embeddings $x_h$, denoted as $\mathcal{F}(x)\label{eq:F}$, where $h$ is the embedding size; an encoder from output labels $y$ to label embeddings $y_h$ as $\mathcal{E}(y)$; a decoder from label embeddings $y_h$ to output labels $y'$, written as $\mathcal{D}(y_h)$. The proposed model is built on two assumptions: first, each instance can be represented from two different aspects, features $x$ and labels $y$, so there exists a common latent space between $x$ and $y$; second, labels can be reproduced by an autoencoder. Based on these two assumptions, we design the object function as below:
\begin{align}
    \mathcal{L}(D) &= \min_{\mathcal{F}, \mathcal{E}, \mathcal{D}} \mathcal{L}_h(x_h, y_h) + \lambda \mathcal{L}_{ae}(y, y') \label{eq:train_loss}
\end{align}
wherein loss $\mathcal{L}_h(x_h, y_h)$ aims to find the common latent space for input $x$ and output $y$ and $\mathcal{L}_{ae}(y, y')$ enforces the output to be reproducible. 
$\lambda$ is a hyper-parameter to balance these two losses. During the training, the model learns a joint network including $\mathcal{F}, \mathcal{E} \text{ and } \mathcal{D}$ to minimize the empirical loss~Eq~(\ref{eq:train_loss}).

During inference, a given input $\hat{x}$ will be first transformed into a vector in latent space $\hat{x}_h = \mathcal{F}(\hat{x})$, which will then be fed into the label decoder to compute the predictions $\hat{y} = \mathcal{D}(\hat{x}_h)$. It is worth mentioning that although the label encoder $\mathcal{E}$ is ignored during the prediction, it is able to exploit cross-label dependency during the label embedding stage~\cite{yeh2017learning}. Recent work also shows that using co-occurring labels information to initialize the neural network can further improve accuracy in multi-label classification~\cite{kurata2016improved,baker2017initializing}. 

\subsection{$\mathcal{L}_h$ and $\mathcal{L}_{ae}$ Loss Functions}\label{sec:loss} 
\textbf{Learning Common Embedding ($\mathcal{L}_{h}$)}. Minimizing the common hidden space loss $\mathcal{L}_h$ has been proposed based on different considerations~\cite{zhang2011multi,yeh2017learning,shen2018compact}, ranging from canonical correlation analysis to alignment of two spaces with a perspective of cross-view. Since the hidden space is usually small and requires less computational cost, we simply employ the mean squared loss for $\mathcal{L}_h$. 

\noindent
\textbf{Reconstructing Output ($\mathcal{L}_{ae}$)}. Unlike $\mathcal{L}_{h}$ with small space, $\mathcal{L}_{ae}$ loss usually involves a large number of labels. Moreover, $\mathcal{L}_{ae}$ also directly affects the classification performance significantly since different loss functions lead to their own properties~\cite{hajiabadi2017extending}. Accordingly, solving such problems with large scale and desirable properties presents open challenges in three aspects: 1) how to improve time efficiency, 2) how to produce comparable labels scores and 3) how to deal with noise labels. Unfortunately, most of the related deep learning methods only target one or two aspects. C2AE attempts to minimize the number of misclassified pairs between relevant and irrelevant labels,
as a result its computational complexity is quadratic with number of labels in the worst case; also it fails to scale well on large number of input features or labels due to its inefficient implementation\footnote{\url{https://github.com/dhruvramani/C2AE-Multilabel-Classification}}.
XML-CNN~\cite{liu2017deep} achieves computational efficiency by training a deep neural network with hidden layers much smaller than the output layer with binary cross-entropy loss (BCE), which has linear complexity in number of labels. Despite this, BCE loss could neither capture label dependencies nor produce directly comparable label scores, since each label is treated independently. Moreover, BCE loss tends to be sensitive to label noise, which is frequently observed in XML data~\cite{reed2014training,ghosh2017robust}. 

To void the aforementioned issues, we propose a marginal-based ranking loss in AutoEncoder: 
{
\begin{align}
    \mathcal{L}_{ae}(y, y') &= \mathcal{L}_P(y, y') + \mathcal{L}_N(y, y') \label{eq:loss}  \\
     \mathcal{L}_P(y, y')   &= \sum_{n\in N(y)} \max_{p \in P(y)} (m + y'_{n} - y'_{p})_{+} \label{eq:p_loss} \\
     \mathcal{L}_N(y, y') &= \sum_{p\in P(y)} \max_{n \in N(y)} (m + y'_{n} - y'_{p})_{+}\label{eq:n_loss}
\end{align}
}
wherein $N(y)$ is the set of negative label indexes, $P(y)$ is the complement of $N(y)$, and margin $m \in [0,1]$ is a hyper-parameter for controlling the minimal distance between positive and negative labels. 
The loss consists of two parts: 1) $\mathcal{L}_P$ targets to raise the minimal score from positive labels over all negative labels at least by $m$; 2) $\mathcal{L}_N$ aims to penalize the most violated negative label under all positive labels by $m$. The proposed loss has the following attractive properties: 1) having linear complexity in number of labels $\mathcal{O}(L)$; 2) capturing the relative rankings between positive and negative labels;
3) tolerating the noisy labels with a tunable hyper-parameter $m$. To explain the last property, assume $y_n'$ and $y_p'$ are the predicted probabilities bounded in $[0,1]$, then on one extreme case with noise-free labels and $m=1$, all positive labels are raised to probability $1$, while negatives are penalized to $0$; on the other extreme case with all random labels, e.g. from i.i.d. Bernoulli distribution, setting $m=0$ indicates that the annotated labels are completely random noises.

\subsection{Dual Attention}\label{sec:attn}
Extracting rich feature representations in XML is helpful for predicting more complicated labels structures, but on the other hand, requires an efficient and feasible method. A recent work (CBAM)~\cite{woo2018cbam} proposes a block attention module, with a Channel-Attention and a Spatial Attention for images tasks only, wherein Channel-Attention emphasizes information from channels, e.g.\ RBG, and Spatial-Attention pays attention to partial areas in an image. By sequentially applying channel and spatial attentions, CBAM is shown to be effective in images classification and object detection. We take advantage of the attentions in CBAM and apply it on textual data.

\begin{figure}[t]
\centering
\includegraphics[width=1.\columnwidth]{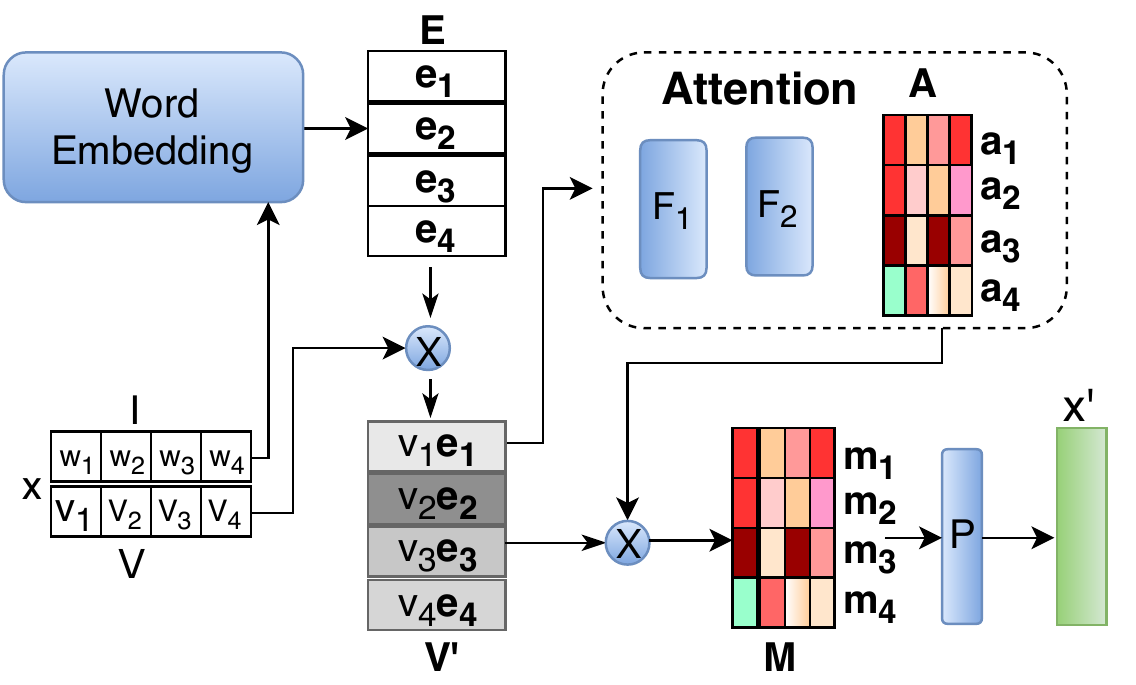}
\caption{Dual-Attention in Feature Embedding.}
\label{fig:ae_att}
\end{figure}

In our proposed attention module, it also consists of spatial-wise and channel-wise attentions. First, we force spatial-wise attention to attend on a list of important words in a way that simply multiply word embeddings by term-frequency or tf-idf (whichever is provided in the feature matrix). It is worth noting that spatial-wise module does not involve any parameters, but it efficiently captures the importance of words with numerical statistics, like tf-idf. 
We demonstrate spatial-wise attention on the left side of Figure~\ref{fig:ae_att}, where the input $x=(I,V)$ contains bag-of-words vector $I=(w_1,w_2,\dots,w_n)^T\in \mathbb{R}^n$ and tf-idf vector $V=(v_1,v_2,\dots,v_n)^T\in \mathbb{R}^n$. Bag-of-words $I$ are fed into an embedding layer $\mathbf{E}=(\mathbf{e}_1,\mathbf{e}_2,\dots,\mathbf{e}_n)^T \in \mathbb{R}^{n \times C}$ to get the word embeddings, where $\mathbf{e}_j \in \mathbb{R}^C$ is word embedding vector of  $w_j$. Then we multiply word embeddings by $V$ to obtain weighted word embeddings: $\mathbf{V}' = (v_1\mathbf{e}_1,v_2\mathbf{e}_2,\dots,v_n\mathbf{e}_n)^T \in \mathbb{R}^{n \times C}$.

The channel attention is designed to emphasize the significant aspects by assigning different weights on bits in a word embedding. For example, in the word embedding of ``apple'', some of the bits may reflect fruit, while others may indicate the company name. To achieve this, we adopt the excitation network from the SENet~\cite{hu2017squeeze} 
with a slight increase in model complexity. The excitation network includes two fully connected layers with a non-linearity activation function in between, see the top-right part in Figure~\ref{fig:ae_att}:

\begin{align}
    \mathbf{A}^T &= \sigma\left( F_2 \delta (F_1 \mathbf{V}'^T) \right) = (\mathbf{a}_1, \mathbf{a}_2,\dots, \mathbf{a}_n)
\end{align}

wherein $\mathbf{A} \in \mathbb{R}^{n \times C}$, $\delta$ and $\sigma$ refer to the two activation functions, $\textit{ReLU}$ and $\textit{Sigmoid}$, $F_1 \in \mathbb{R}^{\frac{C}{r} \times C}$ and $F_2  \in \mathbb{R}^{C \times \frac{C}{r}}$ are the two fully connected layers, with word embedding size  $C$ and reduction ratio $r$. After obtaining the attention matrix $\mathbf{A}$, we can apply those attentions to the weighted word embeddings to get a re-scaled word embedding matrix $\mathbf{M}$: $\mathbf{M}=\mathbf{V}' \circ \mathbf{A} = (\mathbf{m}_1,\mathbf{m}_2,\dots,\mathbf{m}_n)^T \in \mathbb{R}^{n \times C}$, which is computed via the element-wise product. The obtained attention matrix $\mathbf{A}$ introduces dynamics conditioned on the input weighted word embeddings and further boosts feature discriminability. The last step is to feed the re-scaled embedding matrix into an average pooling to obtain the feature embedding $x' \in \mathbb{R}^C$. With the proposed spatial-wise and channel-wise attentions, Rank-AE can learn rich feature representations in an efficient way.

\section{Experiments \& Analysis}
\begin{table}[t]
\begin{centering}
\resizebox{1.0\columnwidth}{!}{%
\begin{tabular}{|c|r|r|r|r|r|r|}
\hline
\textbf{Dataset}   & {\textbf{train}} & {\textbf{test}} & {\textbf{label}} & {\textbf{feature}} & {\textbf{cardinality}} & {\textbf{inst/L}} \\
\hline
\textbf{Delicious*} & 12,920                              & 3,185                              & 983                                & 500                                  & 19.03                                    & 312                                     \\
\textbf{Mediamill*} & 30,993                              & 12,914                             & 101                                & 120                                  & 4.38                                     & 1,902                                    \\
\textbf{RCV*}       & 623,847                             & 155,962                            & 2,456                               & 47,236                                & 4.79                                     & 1,219                                    \\
\textbf{IMDb}      & 27,417                              & 6,740                              & 28                                 & 115,554                               & 2.4                                      & 2,937                                    \\
\textbf{EURLex}    & 15,539                              & 3,909                              & 3,993                               & 5,000                                 & 5.31                                     & 26                                      \\
\textbf{Wiki10}    & 14,146                              & 6,616                              & 30,938                              & 101,938                               & 18.64                                    & 9  
\\ \hline     
\end{tabular}}
\caption{Dataset Characteristics: \textbf{train}, \textbf{test}, \textbf{label} and \textbf{feature} are the numbers of training, testing, labels and features respectively; \textbf{cardinality} is the average number of label per instance; \textbf{inst$/$L} is the average number of instances per label. * indicates only feature matrix is provided and no raw document.}
\label{tab:dataset}
\end{centering}
\end{table}
\subsection{Dataset \& Experiment Setup}
\textbf{Dataset}. Our experiments are conducted on six extreme multi-label datasets and their characteristics are shown in Table~\ref{tab:dataset}, among which IMDb is crawled from online movie database\footnote{\url{{https://www.imdb.com/}}} and the rest five datasets are downloaded from the extreme classification repository\footnote{\url{http://manikvarma.org/downloads/XC/XMLRepository.html}}. For datasets from the repository, we adopt the provided train/test split, and for IMDb we randomly choose $20\%$ of the data as test set and the rest of $80\%$ as training set. For all datasets, we reserve another $20\%$ of training data as validation for tuning hyper-parameters. After tuning, all models are trained on the entire training set.

Among these datasets, three of them are only provided with BoW feature matrix: Delicious, Mediamill (dense feature matrix extracted from image data) and RCV, which are only feasible for the non-deep learning methods (SLEEC, FastXML, PDSparse) and Rank-AE. We provide both feature matrix and raw documents for IMDb, EURLex and Wiki10, which are feasible for both deep learning and non-deep learning methods. For those data with both formats, we remove the words from the raw documents that do not have corresponding BoW features so that the vocabulary size is the same for both deep and non-deep learning methods.

\noindent
\textbf{Evaluation Metrics}. To evaluate the performances of each model, we adopt the metrics that have been widely used in XML: Precision at top k ($P@k$), and the Normalized Discounted Cummulated Gains at top k ($n@k$)~\cite{bhatia2015sparse,prabhu2014fastxml,yen2016pd,liu2017deep}. $P@k$ is a measure based on the fraction of correct predictions in the top k predicted scoring labels and $n@k$ is a normalized metric for Discounted Cumulative Gain:
\begin{align}
    P@K &= \frac{1}{k} \sum_{l \in rank_k(\hat{y})} y_l \\
    DCG@K &= \sum_{l \in rank_k(\hat{y})} \frac{y_l}{log(l+1)} \\
    nDCG@K &= \frac{DCG@K}{\sum_{l=1}^{\min(k,|y|)} \frac{1}{\log(l+1)}}
\end{align}
wherein the $rank_k$ returns $k$ largest indices of the prediction $\hat{y}$ in a descending order, and $|y|$ is the number of positive labels in ground truth. In the results, we report the average $P@k$ and $n@k$ on testing set with $k=1,3,5$ respectively.

\noindent
\textbf{Hyper-parameters}. In Rank-AE, we use the fixed neural network architecture, with two fully connected layers in both \textbf{Encoder} and \textbf{Decoder}, and one fully connected layer following \textbf{Embedding \& Atten} network in \textbf{Feature Embedding}. We also fix most of the hyper-parameters, including hidden dimension $h$ (100 for small number of labels data and 200 for large ones), word embedding size $C=100$, and reduction ratio $r=4$.
The remaining hyper-parameters, such as balance $\lambda$ between $\mathcal{L}_h$ and $\mathcal{L}_{ae}$, margin $m$ in  $\mathcal{L}_{ae}$, and others (decay, learning rate) in the optimization algorithms, are tuned on validation set. In addition, if the vocabulary for BoW is available, e.g.\ IMDb and Wiki10, the \textbf{Word Embedding} component is initialized by Glove\footnote{\url{https://nlp.stanford.edu/projects/glove/}}, a pre-trained word embeddings of 100 dimensions; if it is not, e.g.\ Mediamill, Delicious and RCV, a random initialization is employed. 

For the existing methods with the same train/test split, we take the scores from the original papers for SLEEC, FastXML and PD-Sparse directly. For the new datasets and splits, the hyper-parameters are tuned on the validation set for all methods, as suggested in their papers.

\begin{table*}[t]
\begin{center}
\resizebox{1.8\columnwidth}{!}{%
\begin{tabular}{ccccccccc}
\hline
\textbf{Dataset}                    & \textbf{Metrics} & \textbf{SLEEC} & \textbf{FastXML} & \textbf{PD-Sparse} & \textbf{FastText} & \textbf{Bow-CNN} & \textbf{XML-CNN} & \textbf{Rank-AE} \\
\hline
\multirow{3}{*}{\textbf{Delicious}} & P@1              & 67.59          & \textbf{69.61}   & 51.82              & -                 & -                & -                & 69.26           \\
                                    & P@3              & 61.38          & \textbf{64.12}   & 44.18              & -                 & -                & -                & 62.72           \\
                                    & P@5              & 56.56          & \textbf{59.27}   & 38.95              & -                 & -                & -                & 57.63           \\
\hline
\multirow{3}{*}{\textbf{Mediamill}} & P@1              & \textbf{87.82} & 84.22            & 81.86              & -                 & -                & -                & 86.53           \\
                                    & P@3              & \textbf{73.45} & 67.33            & 62.52              & -                 & -                & -                & 70.17           \\
                                    & P@5              & \textbf{59.17} & 53.04            & 45.11              & -                 & -                & -                & 55.44           \\
\hline
\multirow{3}{*}{\textbf{RCV}}       & P@1              & 90.25          & \textbf{91.23}   & 90.08              & -                 & -                & -                & 90.9            \\
                                    & P@3              & 72.42          & \textbf{73.51}   & 72.03              & -                 & -                & -                & 72.82           \\
                                    & P@5              & 51.88          & \textbf{53.31}   & 51.09              & -                 & -                & -                & 52.05           \\
\hline
\multirow{3}{*}{\textbf{IMDb}}      & P@1              & 51.37          & 66.45            & 66.84              & 69.55             & 66.59            & 75.55            & \textbf{75.91}  \\
                                    & P@3              & 34.46          & 48.32            & 46.29              & 48.76             & 48.42            & 52.59            & \textbf{52.66}  \\
                                    & P@5              & 27.34          & 36.28            & 35.04              & 36.53             & 36.56            & \textbf{38.90}   & 38.48           \\
\hline
\multirow{3}{*}{\textbf{EURLex}}    & P@1              & 79.26          & 71.36            & 76.43              & 71.51             & 64.99            & 76.38            & \textbf{79.52}  \\
                                    & P@3              & 64.30          & 59.90            & 60.37              & 60.37             & 51.68            & 62.81            & \textbf{65.14}  \\
                                    & P@5              & 52.33          & 50.39            & 49.72              & 50.41             & 42.32            & 51.41            & \textbf{53.18}  \\
\hline
\multirow{3}{*}{\textbf{Wiki10}}    & P@1              & \textbf{85.88} & 83.03            & 81.03              & 68.86             & 81.16            & 84.11            & 83.6            \\
                                    & P@3              & \textbf{72.98} & 67.47            & 57.36              & 54.65             & 50.67            & 70.24            & 72.07           \\
                                    & P@5              & \textbf{62.70} & 57.76            & 44.10              & 47.61             & 36.03            & 59.87            & 62.07           \\
\hline
\hline
\textbf{Rank Score}                 & avg              & 2.83           & 3.33             & 4.56               & 4.56              & 5.78             & 2.56             & \textbf{1.78}  
\\ \hline
\end{tabular}
}
\vspace{0.4mm} \\
\small{Table (a): $P@k$: Precision at top k and average ranking scores for different methods.}\\
\vspace{2.mm}
\resizebox{1.8\columnwidth}{!}{%
\begin{tabular}{ccccccccc}
\hline
\textbf{Dataset}                    & \textbf{Metrics} & \textbf{SLEEC} & \textbf{FastXML} & \textbf{PD-Sparse} & \textbf{FastText} & \textbf{Bow-CNN} & \textbf{XML-CNN} & \textbf{Rank-AE} \\ \hline
\multirow{3}{*}{\textbf{Delicious}} & n@1              & 67.59          & \textbf{69.61}   & 51.82              & -                 & -                & -                & 69.26           \\
                                    & n@3              & 62.87          & \textbf{65.47}   & 46.00              & -                 & -                & -                & 64.16           \\
                                    & n@5              & 59.28          & \textbf{61.90}   & 42.02              & -                 & -                & -                & 60.39           \\
\hline
\multirow{3}{*}{\textbf{Mediamill}} & n@1              & \textbf{87.82} & 84.22            & 81.86              & -                 & -                & -                & 86.53           \\
                                    & n@3              & \textbf{81.50} & 75.41            & 70.21              & -                 & -                & -                & 78.36           \\
                                    & n@5              & \textbf{79.22} & 72.37            & 63.71              & -                 & -                & -                & 75.28           \\
\hline
\multirow{3}{*}{\textbf{RCV}}       & n@1              & 90.25          & \textbf{91.23}   & 90.08              & -                 & -                & -                & 90.9            \\
                                    & n@3              & 88.86          & \textbf{89.63}   & 88.50              & -                 & -                & -                & 89.29           \\
                                    & n@5              & 89.49          & \textbf{90.33}   & 88.79              & -                 & -                & -                & 89.75           \\
\hline
\multirow{3}{*}{\textbf{IMDb}}      & n@1              & 51.37          & 66.45            & 66.84              & 69.55             & 66.59            & 75.55            & \textbf{75.91}  \\
                                    & n@3              & 49.75          & 67.14            & 64.84              & 68.47             & 67.26            & \textbf{74.02}   & 73.5            \\
                                    & n@5              & 54.43          & 71.72            & 69.69              & 72.99             & 72.07            & \textbf{78.48}   & 77.37           \\
\hline
\multirow{3}{*}{\textbf{EURLex}}    & n@1              & 79.26          & 71.36            & 76.43              & 71.51             & 64.99            & 76.38            & \textbf{79.52}  \\
                                    & n@3              & 68.13          & 62.87            & 64.31              & 63.32             & 55.03            & 66.28            & \textbf{68.76}  \\
                                    & n@5              & 61.60          & 58.06            & 58.78              & 58.56             & 49.92            & 60.32            & \textbf{62.33}  \\
\hline
\multirow{3}{*}{\textbf{Wiki10}}    & n@1              & \textbf{85.88} & 83.03            & 81.03              & 68.86             & 81.16            & 84.11            & 83.6            \\
                                    & n@3              & \textbf{76.02} & 75.35            & 62.62              & 56.72             & 56.14            & 73.52            & 74.78           \\
                                    & n@5              & \textbf{68.13} & 63.36            & 52.03              & 51.19             & 45.29            & 65.50            & 67.18           \\
\hline
\hline
\textbf{Rank Score}                 & \textbf{avg}     & 2.83           & 3.22             & 4.33               & 4.67              & 5.78             & 2.56             & \textbf{1.89}  
\\ \hline
\end{tabular}
}
\vspace{0.4mm} \\
{Table (b): $n@k$: nDCG at top k and average ranking scores  for different methods.}\\
\vspace{2.mm}
\end{center}
\caption{Comparisons with other methods. '-' indicates unavailable due to raw documents are not available for these deep learning methods, and number in \textbf{bold} is the best result in the line.}
\label{tab:result}
\end{table*}
\subsection{Comparisons with Related Methods}
We evaluate the proposed Rank-AE with other six state-of-the-art methods, SLEEC, FastXML, PD-Sparse, FastText, Bow-CNN and XML-CNN, which are the leading methods among their categories. Among them, FastText, Bow-CNN and XML-CNN only take raw documents, which are not available for Delicious, Mediamill and RCV datasets. For Rank-AE, we adopt the raw text as the input for IMDb, and feature matrix for the rest.

The performances evaluated on $P@k$ and $n@k$ with $k=1,3,5$ are summarized in Table \ref{tab:result} (a) and (b) separately. As reported, Rank-AE reaches the best performances on two datasets (IMDb and EURLex) out of 6 datasets, while SLEEC achieves the best performances on Mediamill and Wiki10, and FastXML performs the best on Delicious and RCV. In general, SLEEC and FastXML are very competitive to each other in non-deep learning methods, but PD-Sparse performs worse. Rank-AE always performs better than PD-Sparse with at least $1\%$ increase, up to almost $20\%$ improvement on Delicious data. When compared with FastXML, Rank-AE outperforms on 4 datasets with $1\%$ to $10\%$ growth, but underperforms on Delicious and RCV with $1\%$ decrease. SLEEC, as the best non-deep learning method in our experiments, performs almost identical to Rank-AE, but on IMDb data, it performs $7\% \sim 15\%$ less than non-deep methods, and even worse than Rank-AE.

Comparing Rank-AE with deep learning methods, we narrow down to three datasets with available raw documents: IMDb, EURLex and Wiki10. As shown in Table~\ref{tab:result}, FastText and Bow-CNN, not planned for XML but for multi-class, perform much worse than XML-CNN and Rank-AE as expected. On the other hand, XML-CNN achieves close performance to Rank-AE: with similar performance on IMDb dataset, but lower scores on EURLex and Wiki10 with $2\%$ drop in $P@k$ and $n@k$. In spite of this, Rank-AE, trained on feature matrix for EURLex and Wiki10, surprisingly performs better than XML-CNN on raw data. 

In the comparisons, there is no such method that could perform the best on all datasets. 
We discover that each dataset has its own intrinsic properties, such as diversity of labels, number of features, average number of relevant labels per instance and average number of training instances per label, see Table~\ref{tab:dataset}. All those properties will affect training procedure, for example, how much flexibility a model should be in order to explain labels well by the given training data. Because those factors are always changing from data to data, they also influence the performances on different models. In order to have a reasonable comparisons, we report the average ranking score for each method. To compute the average ranks, we first rank the methods based on their performance in each row in Table (a) and Table (b), then average them through all rows, and report the final ranking scores in the last row of each table. The average ranking scores show that Rank-AE is the best model with ranking scores $1.78$ in $P@k$ and $1.89$ in $n@k$.

\subsection{Comparisons with Noise Labels}
As mentioned previously, noisy labels in XML are a quite common issue in the real-world applications~\cite{yeh2017learning,ghosh2017robust}, but our proposed marginal ranking loss naturally mitigates this problem. Since IMDb is a real-world dataset with relatively clean labels, we conduct the noise experiments on it. In the experiments, we control the noise labels in two different ways: 1) missing labels: changing each positive label from $y_l=1$ to $y_l=0$ with certain rate, 2) both missing and invalid labels: flipping either from positive to negative or from negative to positive with a noise rate. The noise rates are varied from $0\%$ to $60\%$ on $80\%$ of the training set, and the rest of $20\%$ is noise-free validation set for model selection. We select five algorithms: FastXML, PD-Sparse, XML-CNN, Rank-AE and BCE-AE, wherein BCE-AE is our proposed method but using binary cross-entropy loss in $\mathcal{L}_{ae}(y, y')$. Comparing BCE-AE with Rank-AE can be used to verify whether the robustness to label noise is due to the use of marginal ranking loss.

\begin{figure}[t] 
  \subfigure[Missing Rate]{%
    \includegraphics[width=1\columnwidth]{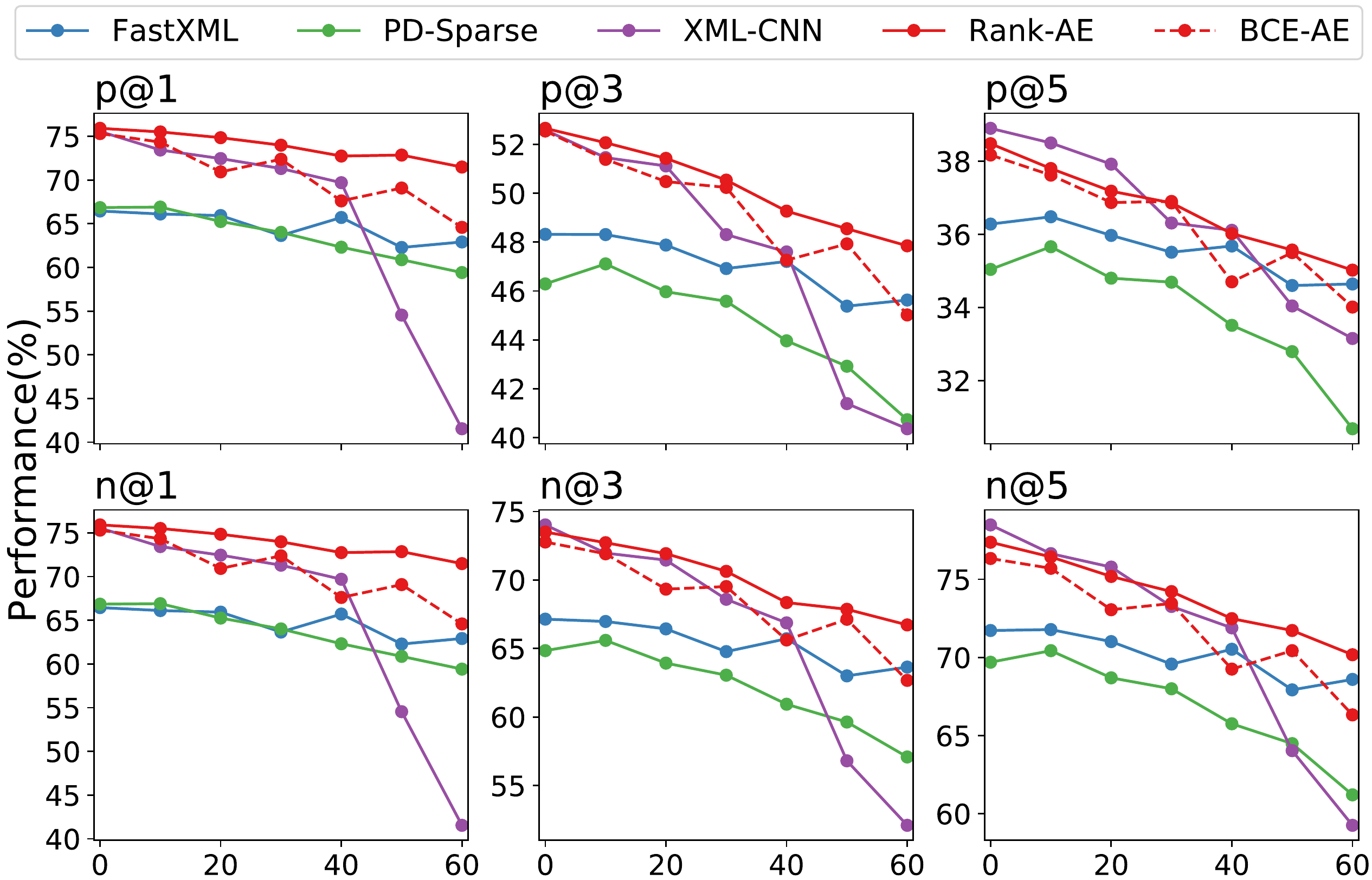} \label{fig:noise_1} 
  }
  \subfigure[Missing and Invalid Rate]{%
    \includegraphics[width=1\columnwidth]{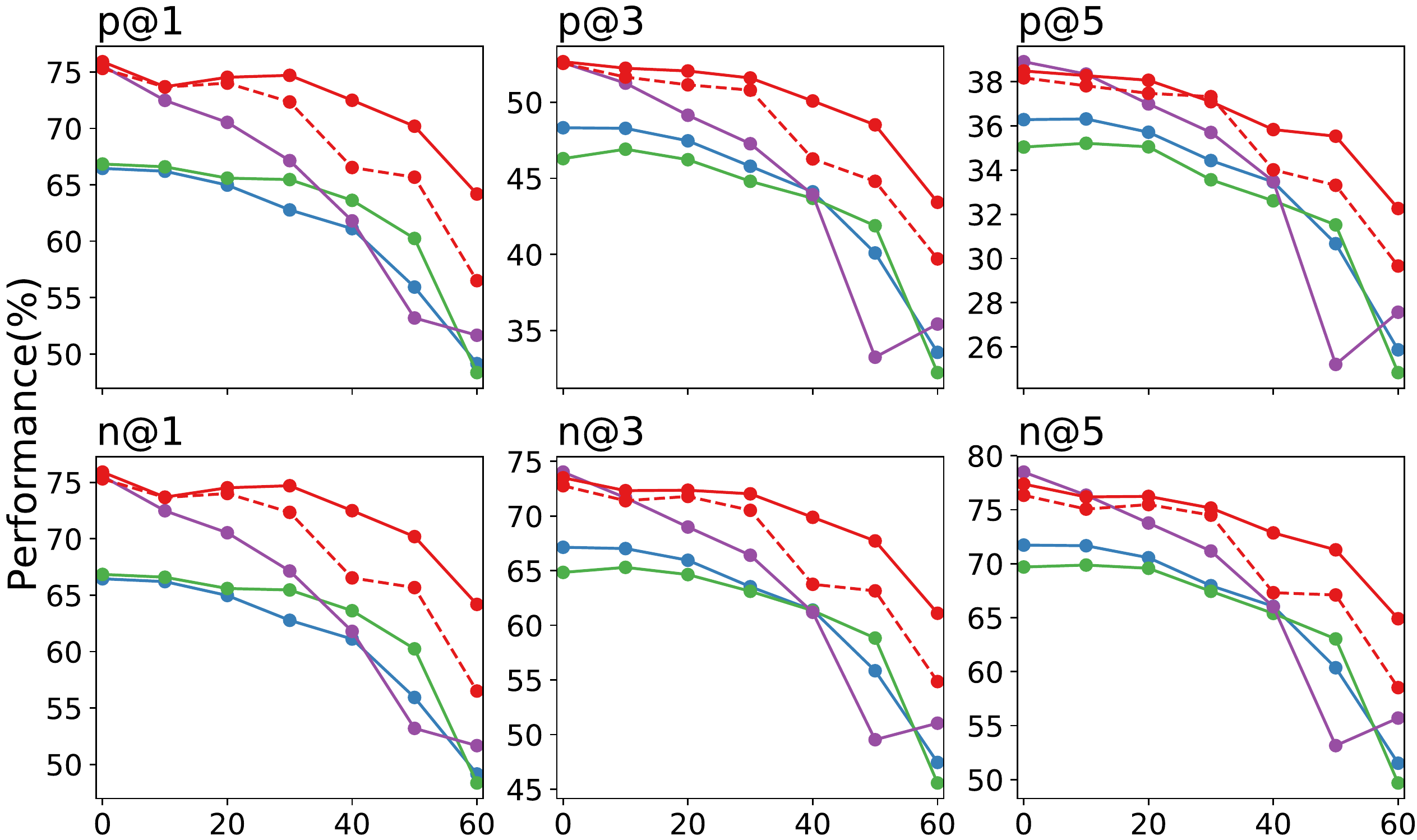} \label{fig:noise_2} 
  } 
  \caption{Comparisons on noisy labelling IMDb data.} \label{fig:noise}
\end{figure}

The performances are reported on the same clean test set, shown in Figure~\ref{fig:noise}. Rank-AE consistently outperforms other four approaches and has the best robustness tolerating noise labels. Besides, FastXML and PD-Sparse are more tolerant to missing noises than XML-CNN, which may due to XML-CNN has greater capacity and thus more prone to over-fitting the noise. Furthermore, when comparing Rank-AE with BCE-AE, both of which share the same structure but have different loss functions, the proposed marginal-based ranking loss seems to be robuster than binary cross-entropy loss.


\begin{figure}[t]
\centering
\includegraphics[width=1.0\columnwidth]{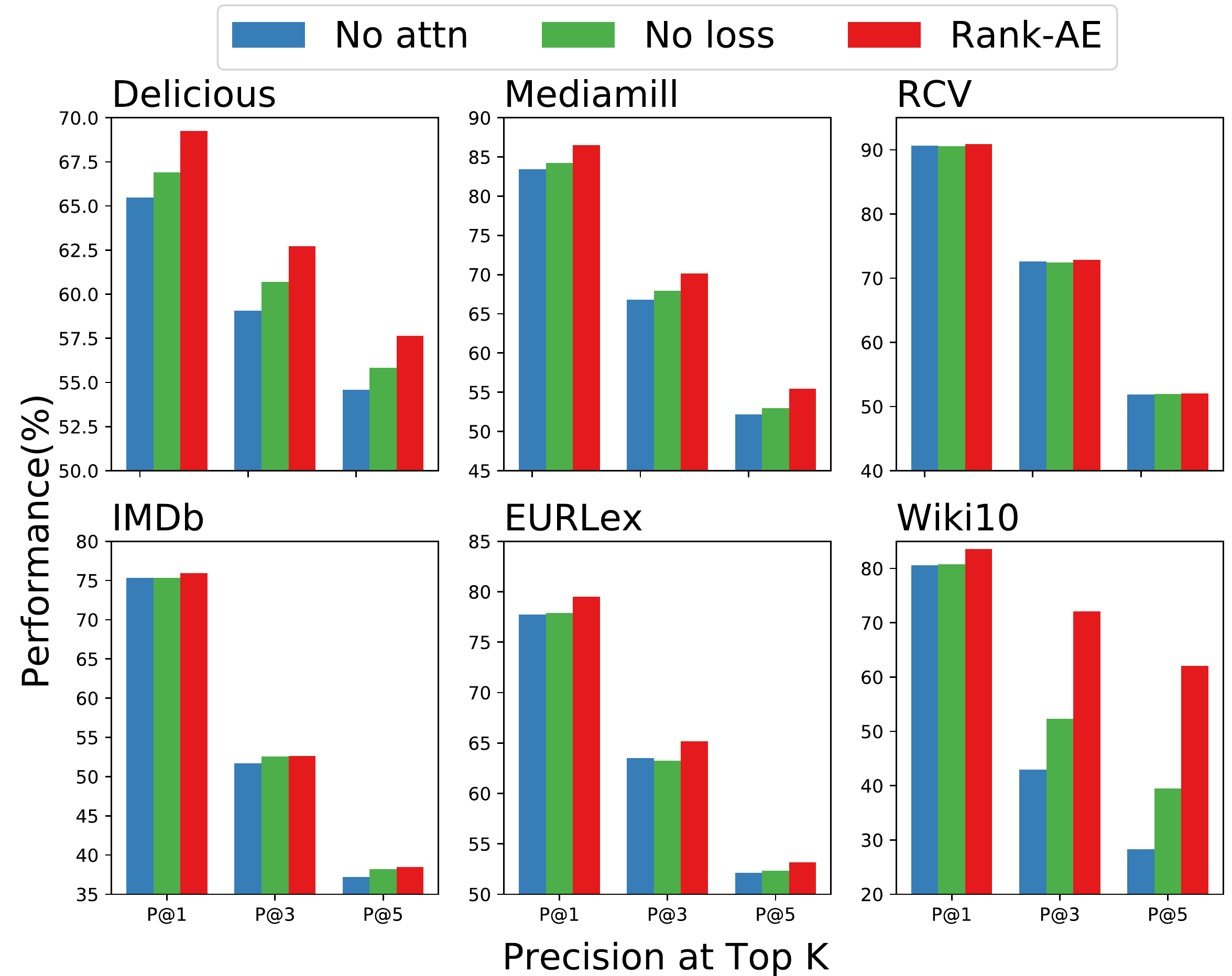}
\caption{Precision at top k comparisons: No attn is no attention Rank-AE; No loss is using binary cross entropy lose instead of marginal ranking loss; Rank-AE is our proposed model.}
\label{fig:prec_comp}
\end{figure}

\begin{figure*}[ht]
\centering
\includegraphics[width=1.6\columnwidth]{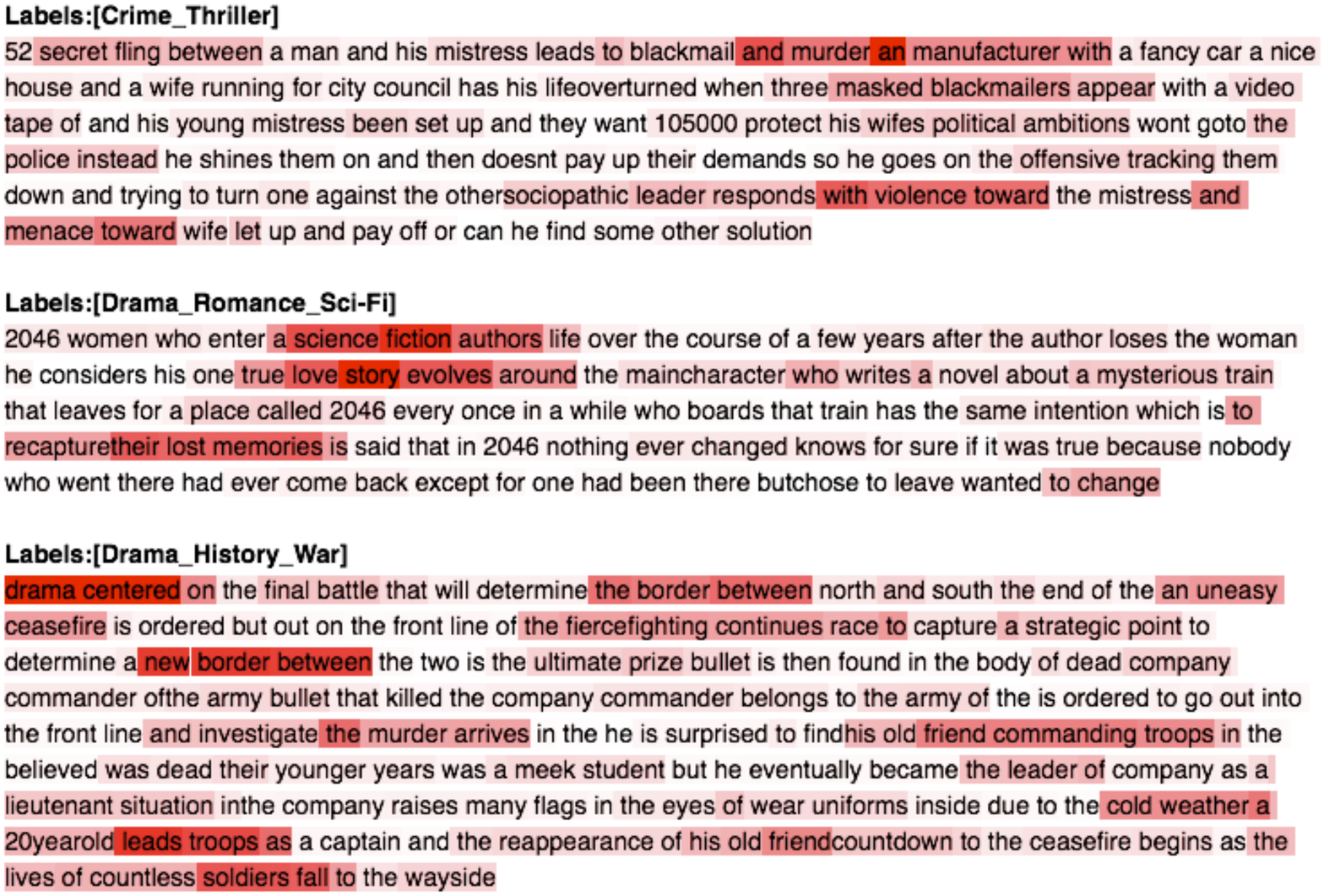}
\caption{Visualization for Attention in Rank-AE. The labels are movie genres split by underline, and the input is a movie story line.}.
\label{fig:movie_genres}
\end{figure*}

\subsection{More Analysis in Rank-AE}
\textbf{Ablation Study}. 
The effectiveness and robustness of Rank-AE have been demonstrated in the previous section. However, it is not clear to us yet that if the effectiveness benefits from the proposed components, such as attention mechanism and marginal ranking loss. To further understand the impacts from these two factors, we conduct a controlling experiment with three different settings: 1) removing the Attention component $\mathcal{A}$ in Figure~\ref{fig:ae_att} from Rank-AE, in which case $\mathbf{V}'$ is directly passed to the average pooling to obtain $x'$, called \textbf{No attn}; or 2) examining the performances by replacing the marginal ranking loss ($\mathcal{L}_{ae}$) with a binary cross entropy loss, named \textbf{No loss}; or 3) keeping the original \textbf{Rank-AE} without any change. In Figure~\ref{fig:prec_comp}, $P@k$ is reported on the six datasets for the ablation experiment, because $n@k$ is similar to $P@k$, thus eliminated here. The comparisons results show that Rank-AE without any change works better than the other two on all datasets consistently, especially on Wiki10. First, channel-attention extracts richer information from the word embeddings by introducing the channel weights. Thus, it is more suitable when classification tasks become more complicated and a word more likely represents multiple aspects. Second, Rank-AE gains some advantage of tolerating noise labels with marginal ranking loss comparing to BCE loss. We could even further infer that IMDb and RCV may have relatively less noise labels since the performance does not benefit much from the marginal ranking loss.

\noindent
\textbf{Channel-Attention Visualization}. Our channel-attention is implemented by an excitation network, which is adopted from SENet~\cite{hu2017squeeze} and only applied to images before. To demonstrate its effectiveness and feasibility on textual data, we employ the visualization tool~\cite{lin2017structured} to highlight important words based on the attention output. Specifically, we run our method on IMDb dataset, wherein each instance is a movie story associated with relevant genres as labels. Instead of extracting $\mathbf{V}'$ matrix using the proposed spatial-wise attention, we obtain a fixed size embeddings from a bidirectional LSTM on variable length of sentence, fed to our channel-attention network. Through the channel-attention network, we can observe the attention matrix $\mathbf{A}$ for each input document. By summing up the attention weights of each word embedding vector, we can visualize the overall attention for that word with the visualization tool\footnote{The visualization tool is provided by \\ \url{https://github.com/kaushalshetty/Structured-Self-Attention}}. We randomly select three movies from IMDb testing set (See Figure~\ref{fig:movie_genres}). By looking at the highlighted regions, we can see that the proposed channel-attention is able to focus more on the words that are highly related to the topics.

\section{Conclusion}
In this paper, we propose a marginal ranking loss, which not only predicts comparable labels scores between labels, more suitable for ranking metrics, but also consistently performs better on noisy labeling data, with both missing and invalid labels. In addition, the dual-attention component allows Rank-AE to learn more powerful feature representations efficiently. By integrating those components, Rank-AE usually achieves the best or the second best on six benchmark XML datasets comparing with other outstanding methods in state-of-the-art.

\section*{Acknowledgements}
Most of this work was done when Bingyu Wang and Wei Sun were interning at JD.Com Inc. We thank Javed Aslam, Virgil Pavlu and Cheng Li from Khoury College of Computer Sciences at Northeastern University for comments that greatly improved the manuscript. We would also like to show our gratitude to our anonymous NAACL-HLT reviwers for the helpful suggestions to make the paper better.




\bibliography{naaclhlt2019}
\bibliographystyle{acl_natbib}

\end{document}